\documentclass[10pt]{article}
\usepackage[letterpaper]{geometry}
\usepackage{hicss}
\usepackage{times}
\usepackage[none]{hyphenat}
\usepackage{url}
\usepackage{latexsym}
\usepackage[frozencache=true,cachedir=minted-cache]{minted} 
\usepackage{indentfirst}
\usepackage{graphicx}
\graphicspath{{images/}}
\usepackage[
    style=apa,
  ]{biblatex}
\usepackage{amsmath,amssymb,amsfonts}
\usepackage{tkz-kiviat} 
\usetikzlibrary{arrows}
\usepackage{algorithm}
\usepackage{color,soul}
\usepackage{algcompatible}
\usepackage{graphicx}
\usepackage{hyperref}
\usepackage{textcomp}
\usepackage{listings,xcolor}
\usepackage{amssymb,amsmath,amsfonts,amsthm} 
\usepackage[noend]{algpseudocode}
\usepackage{mathtools,bm}

\title{The Systems Engineering Approach in Times of\\Large Language Models}

\author{Christian Cabrera \\
  University of Cambridge\\United Kingdom \\
  {\underline{ chc79@cam.ac.uk}} \\ \\
  Jennifer Schooling \\
  Anglia Ruskin University\\United Kingdom \\
  {\underline{ jennifer.schooling@aru.ac.uk} } \\ \And
  Viviana Bastidas \\
  University of Cambridge\\United Kingdom \\
  {\underline{ vab44@cam.ac.uk} } \\ \\
  Neil D. Lawrence\\
  University of Cambridge\\United Kingdom \\
  {\underline{ ndl21@cam.ac.uk} } \\ 
}

\date{}

\begin{document}

\maketitle

\footnotetext[1]{This paper has been accepted for the upcoming 58th Hawaii International Conference on System Sciences (HICSS-58).}

\begin{abstract}
Using Large Language Models (LLMs) to address critical societal problems requires adopting this novel technology into socio-technical systems. However, the complexity of such systems and the nature of LLMs challenge such a vision. It is unlikely that the solution to such challenges will come from the Artificial Intelligence (AI) community itself. Instead, the Systems Engineering approach is better equipped to facilitate the adoption of LLMs by prioritising the problems and their context before any other aspects. This paper introduces the challenges LLMs generate and surveys systems research efforts for engineering AI-based systems. We reveal how the systems engineering principles have supported addressing similar issues to the ones LLMs pose and discuss our findings to provide future directions for adopting LLMs. 
\end{abstract}

\subsubsection*{Keywords:}

Systems Engineering, Socio-technical Systems, AI, and Large Language Models.

\section{Introduction}
\label{sec:introduction}

Large Language Models (LLMs) leverage neural network architectures trained on large amounts of data to learn underlying language patterns. LLMs generate content in formats humans understand based on these architectures~\parencite{feuerriegel2024generative}. Such ability creates novel human-machine interfaces~\parencite{cabrera2024self} for adopting AI at different levels of our society. Generative AI technologies promise new applications for addressing critical problems in diverse domains. 

The complexity of socio-technical systems and the LLMs' nature challenge the realisation of this vision. Social problems have critical requirements that demand reliable systems. LLMs rely on probabilistic models that make systems' components based on these technologies non-deterministic, data-driven, and prone to hallucinations~\parencite{dantonoli2024large} impacting the alignment and reliability of the systems. LLMs operate as black-boxes~\parencite{feuerriegel2024generative}, which impact systems' accountability and interpretability as designers and users do not control and understand the systems (i.e., intellectual debt). Social problems usually appear in resource-constrained environments. Building LLMs is an expensive process that causes significant environmental concerns because it generates an immense carbon footprint~\parencite{schwartz2020greenai}.

The outlined challenges require interdisciplinary research to align societal problems, systems, and AI technical advances. The systems engineering approach is equipped with principles to facilitate this alignment by prioritising the problems and their context before considering the technologies for their resolution. We envisage an ecosystem where systems engineering and LLMs mutually benefit instead of the naive belief that benefits come from the LLMs to the domains in one direction. This paper surveys how researchers have used the system engineering approach to design AI-based systems since 2017 (i.e., when current LLM technologies emerged) as a first step to building such an ecosystem. The main research question we want to answer is \textit{how does current research use the systems engineering approach to address challenges similar to the ones LLMs impose on socio-technical systems?} 
\section{Related Work}
\label{sec:related-work}

Researchers have revised the system engineering concept in light of advances in AI before.~\textcite{sommerville2019artificial} identifies a gap between systems engineering and AI communities, given the different process models these communities use. The systems engineering process starts with requirements and ends with a valid system. AI follows an exploratory process that formulates, develops and evaluates ideas.~\textcite{wade2020vlsirev} review the systems engineering approach for AI inspired by the Very Large Scale Integration (VSLI) revolution. The authors make a parallel between the challenges the transistor technology faced before its adoption and the current challenges that the adoption of AI faces. The common denominator in both cases is that the technologies rapidly surpassed the ability to engineer them.~\textcite{llinas2021review} review the role of Systems Engineering for AI-based systems through time and formulate future research areas. This review focuses on expert systems and machine learning (ML) technologies, which are predecessors of novel generative AI.~\textcite{vanderlinde2022aisechallenges} review the new failure modes that AI introduces into systems and their potential challenges for systems engineering. They claim that traditional systems engineering methods are insufficient to design robust systems.~\textcite{pfrommer2022ki} introduces the concept of AI Systems Engineering as a new branch of Systems Engineering that addresses the systematic development and operation of AI-based solutions as parts of systems that master complex tasks. Related research coincides with the importance of the systems approach for AI-based systems. Some advocate for more systems engineering practices in AI projects, while others claim the systems engineering approach should evolve to handle new technologies. We nurture these discussions by surveying applications of the systems engineering approach from the perspective of the latest AI shift.
\section{Socio-technical Systems and LLMs Challenges}
\label{sec:llm-challenges}

This section summarises the most common issues the applications of LLMs generate.

\subsection{Systems Alignment and Reliability}

Socio-technical systems address problems that affect the interests of multiple stakeholders. Engineers define system requirements according to such problems. Alignment issues emerge when systems do not accomplish their requirements~\parencite{bastidas2022concepts}. 

Alignment issues can emerge when systems integrate LLMs-based components. LLMs are probabilistic and inject uncertainty or even errors (i.e., hallucinations~\parencite{feuerriegel2024generative}) in their outputs. Uncertain, unexpected, or incorrect outputs can go against the satisfaction of systems requirements (i.e., misalignment). In addition, the quality of the LLMs outputs depends on the quality of the data they are trained on~\parencite{cabrera2023real}. If the data is biased and contains unfair relationships, LLMs will produce biased and unfair outputs~\parencite{dantonoli2024large}. Such outputs can then propagate through systems, impacting their performance and how systems satisfy their requirements. These limitations negatively impact system reliability, which in turn causes a lack of trust. End users can perceive systems as unhelpful or harmful, even when systems are well-aligned with their requirements because of such unreliability.
 
\subsection{Interpretability and Accountability}

Socio-technical systems comprise separate but interdependent social and technical subsystems~\parencite{baxter2010socio}. Defining and understanding subsystems' scope, responsibilities, and impact is crucial in designing and deploying systems. Interpretability and accountability issues emerge when there is no clear definition of the responsibilities of systems components and users have difficulty understanding their complexity and behaviour.

Complex and large deep neural network architectures are behind LLMs. These architectures have billions of parameters, and the reasons behind their outputs are often hard to understand. A whole research area explores how to improve the explainability of machine learning models~\parencite{zhao2024expllmsurvey}. LLMs bring the explainability issue to the systems level as they operate as black-box components~\parencite{cabrera2023real}. Integrating LLMs as black boxes in larger systems negatively impacts their interpretability as these generate intellectual debt. Engineers, practitioners, and end-users know that ML-based components can solve challenging problems but do not always know why these components generate their outputs~\parencite{zittrain2022debt}. This lack of understanding also affects systems' accountability and trustworthiness because components' functionalities and boundaries are unclear, which makes it difficult to detect the sources of misbehaviour. 

\subsection{Maintainability and Sustainability}

Socio-technical systems rely on optimising social and technical subsystems~\parencite{baxter2010socio}. This optimisation must consider constraints over natural, human, and economic resources. Engineers must create self-maintaining and sustainable solutions as the planet is in an environmental crisis.

Integrating LLMs into systems requires developing LLMs trained on specific datasets~\parencite{dantonoli2024large}. However, building, fine-tuning, and testing LLMs cause significant environmental concerns as they generate an immense carbon footprint~\parencite{feuerriegel2024generative}. These issues threaten the sustainability of socio-technical systems against the optimal usage of resources. The intellectual debt~\parencite{zittrain2022debt} also challenges self-maintaining socio-technical systems based on LLMs. Self-maintaining systems require an autonomous understanding of the system to identify intervention points and anticipate cascade effects.

\subsection{Security and Privacy}

Socio-technical systems create environments where data flows between external entities, internal actors, subsystems, and their components~\parencite{baxter2010socio}. This data might be sensible depending on the application domain (e.g., health care, urban planning, etc). Engineers must prioritise privacy and security requirements when designing socio-technical systems that manipulate sensible data.

LLMs inject new security threats into critical socio-technical systems. These models create risks of leaking confidential and sensitive information from the model's training data. Recent research shows that adversarial attacks or prompt engineering strategies can extract personal information from LLMs~\parencite{dantonoli2024large}. The complexity, lack of transparency, and limited accountability that LLMs add to the systems challenge the mitigation strategies developed by different communities to address security risks.
\section{Survey Methodology}
\label{sec:methodology}

We survey papers that apply systems engineering approaches to address the described challenges in the context of AI-based systems. Our survey uses a tool\footnote{Survey Tool: \url{https://github.com/cabrerac/semi-automatic-literature-survey/tree/sys-llms-survey}} that semi-automates the paper selection process~\parencite{kitchenham20132049}. We use the following query as an input to search for the relevant papers: ("systems engineering" OR "systems thinking" OR "dependable systems" OR "engineering AI") AND ("generative ai" OR "large language model" OR "llm" OR "artificial intelligence" OR "ai" OR "ml" OR "deep learning"). The parameters file for this search process is here\footnote{Survey parameters file: \url{https://github.com/cabrerac/semi-automatic-literature-survey/blob/sys-llms-survey/parameters_sys.yaml}}. We searched for papers in IEEEXplore, Springer Nature, Scopus, Semantic Scholar, CORE, and ArXiv. The tool retrieved a total of 3,504 papers. We apply a semantic filter comparing the semantic embedding of the titles and abstracts of these papers with a description of the type of papers we are searching for. The semantic filter selected 253 papers. We manually filtered these papers by reading the title and abstract and skimming the full text. Thirty-four papers passed the manual filters. The semi-automatic tool applies a snowballing process that produced 22 papers. We manually selected 5 out of the 22. We filtered thirty-nine papers for full-text reading. We read the final list of papers to perform our analysis. After this process, we selected 24 papers for our report. The code and data of the survey process are available online\footnote{Survey process files: \url{https://github.com/cabrerac/semi-automatic-literature-survey/tree/sys-llms-survey/papers/sys_search/2024_04_07}} for transparency and reproducibility.
\section{Systems Engineering in Times of LLMs}
\label{sec:survey}

\begin{figure}[t!]
    \centering
    \includegraphics[width=0.85\linewidth]{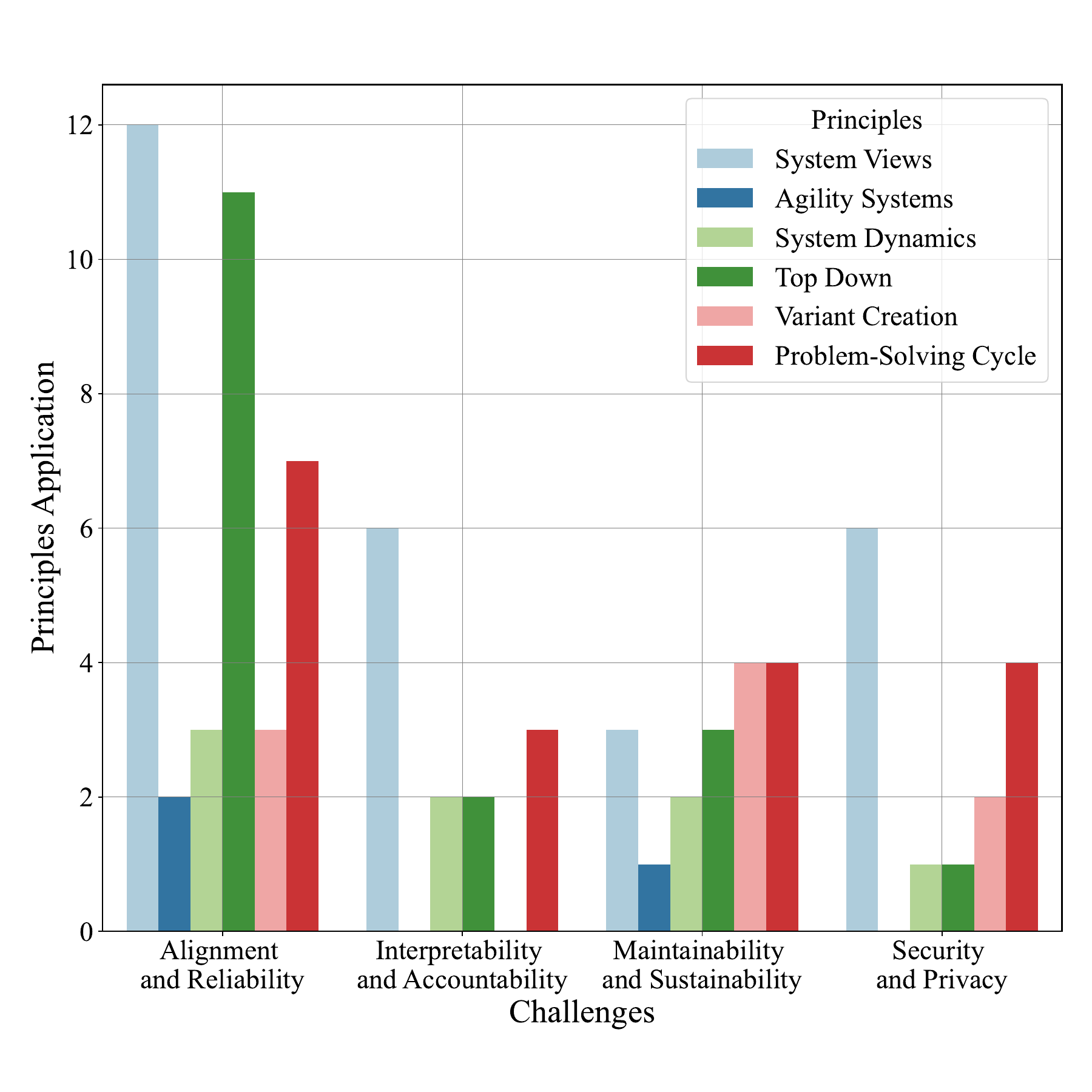}
    \vspace{-5mm}
    \caption{Number of papers that apply a principle to address the identified LLMs Challenges (Section~\ref{sec:llm-challenges}).}
    \label{fig:bar}
    \vspace{-7mm}
\end{figure}

Figure~\ref{fig:bar} presents how the papers applied the systems engineering principles.~\textcite{haberfellner2019systems} classify the principles into two categories. The first category corresponds to \textit{Systems Thinking} and groups the principles of \textit{Systems Views, Agility Systems, and System Dynamics}. The second category corresponds to the \textit{Systems Engineering Process Model} and includes the \textit{Top-Down, Variant Creation, and Problem-Solving Cycle} principles. Our survey shows that current research prioritises \textit{Alignment and Reliability} to leverage AI features while assuring that systems behave as expected. The principles that guide most approaches are the \textit{Systems Views, Top-Down, and Problem-Solving Cycle}. The rest of this section details our findings.

\subsection{Systems Alignment and Reliability}

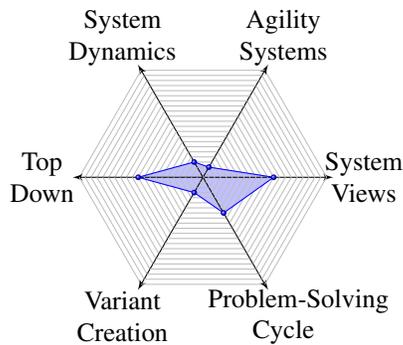
\begin{figure}[t!]
    \centering
    \begin{tikzpicture}[scale=0.18]
        \tkzKiviatDiagram[lattice=21, label space=2.75, gap=.435]{System Views, Agility Systems, System Dynamics, Top\\Down, Variant Creation, Problem-Solving Cycle}
        \tkzKiviatLine[mark=ball, mark size=6pt,color=blue, fill=blue, opacity=.25](12,2,3,11,3,7)
    \end{tikzpicture}
    \vspace{-4mm}
    \caption{Radar chart comparing applied principles in the context of Systems Alignment and Reliability.}
    \label{fig:radar-alignment}
    \vspace{-7mm}
\end{figure}

Around 87\% of papers (i.e., 21 out of 24) address alignment and reliability issues. Figure~\ref{fig:radar-alignment} shows how papers apply the principles. Most works apply the \textit{System Views and Top-Down} principles followed by the \textit{Problem-Solving Cycle} principle. The System Views principle defines systems from different perspectives and facilitates the design of reliable systems that align with their intents.~\textcite{fujii2020guidelines} propose quality guidelines for systems that rely on ML components with limited accuracy. These guidelines establish checkpoints along the system lifecycle, including data integrity, model robustness, system quality, process agility, and customer expectation viewpoints.~\textcite{salwei2022healthcare} propose a model to represent systems that apply AI in healthcare. This model includes people, organisations, tasks, AI technologies, other technologies, and physical environments. The authors use this model to analyse if AI generates positive or negative outcomes. For example, the model can support observing the practitioners' workflow to identify bottlenecks and assess AI tools' usability.~\textcite{wang2022health} propose to model the healthcare systems using three perspectives to support a transparent understanding of the interactions and connections between governance institutions, patients' perceptions, and complex healthcare systems. The digital and governance perspectives improve patients' perceptions when systems prioritise patients' needs and regulation and education policies are in place.

The top-down principle decomposes the problem and the solution to guarantee that the latter addresses the former.~\textcite{meyer2019seagents} uses this principle to align RL solutions with high-level requirements. The misalignment between both sides emerges from the fact that ML model goals (e.g., accuracy) are distant from the requirements of socio-technical systems. The component-based software development model bridges this gap.~\textcite{lavazza2021intensive} propose a notation to represent uncertainty in AI-Intensive Systems (AIIS). This notation decomposes the system into infrastructure, ML components, and traditional software, enabling an alignment analysis.~\textcite{langford2022trusted} introduce a framework to manage learning-enabled components (LECs) for safety-critical tasks. The framework decomposes the uncertainty problem and creates microservices to manage LECs. An adversarial detection microservice identifies when LECs are likely to misbehave to avoid their usage.~\textcite{hedin2023strengthening} propose to support the human component of AI-based systems. They improve systems reliability by creating a better-informed public through AI benchmarks. The paper builds the benchmarks by adopting top-down strategies from high-level requirements to systems components.

Combining principles supports solutions that consider more aspects of the systems.~\textcite{mattioli2023industry} introduce the Confiance.ai framework for verifying and validating industrial systems. The framework formalises ten viewpoints in a metamodel that captures concepts around the system's specification, lifecycle, technical implementation, associated processes, people, governance, and risk and quality management.~\textcite{adedjouma2022dependable} uses these models for an assurance use case. Requirements are decomposed and mapped to the system's components that realise them. Such a mapping enables evidence-based reliability specification.~\textcite{nabavi2022five} propose the Five Ps as an analytical and planning tool that models AI-based systems from the Problem and solution viewpoints and decomposes the system from requirements to components into Purpose, Pathway, Process and Parameter zones. The tool identifies the points in the zones where interventions can generate more value for Responsible AI requirements (i.e., fairness and trustworthiness.).

~\textcite{laato2022governance} propose to close the gap between governance and ML-based systems by considering different viewpoints to define governance concepts and requirements. These concepts are mapped to the development process (i.e., the \textit{Problem-Solving Cycle} principle) to establish a governance model for system design, development, and operation. The system design dimension focuses on the system requirements, data needs, and alignment.~\textcite{zeller2024toward} introduce a workflow for continuous deployment and safety assurance of ML-based systems. This process includes testing and verification steps along the systems' lifecycle, considering three viewpoints to provide solutions that operate according to safety functional requirements.~\textcite{solomonides2021amias} present governance principles and a systems lifecycle for adopting AI into healthcare systems. The authors propose principles from the medical practice (i.e., the Belmont principles), healthcare organisations, and technology perspectives. A process model applies these principles in the inception, development, deployment, maintenance, and decommissioning steps. Systems alignment and reliability rely on rich specifications at inception, rigorous testing at development, and auditing mechanisms and educational efforts at deployment.

\textcite{pedroza2019safe} propose a process based on a reference architecture that models AI systems' mission and principles. This architecture drives a safe-by-design development process that compromises 11 phases that decompose the systems' goals into functional analysis and assessment of AI components.~\cite{yu2024management} proposes five viewpoints: design objectives, system boundaries, system architectures, predictability and emergence, and learning and adaptation. They advocate for more flexible, inclusive, and social-oriented design objectives, more fluid and diffused boundaries between systems' components, decentralised and open architectures that prioritise data, and the integration of predictive methods to respond to emergent behaviours and adapt to changing goals. This work addresses reliability and alignment by allowing generativity (i.e., design objectives, system boundaries, and system architecture dimensions) while assuring criticality (i.e., systems predictability and emergence and systems learning and adaptation dimensions).~\textcite{hasterok2022process} present a process model for non-deterministic AI-based systems. In contrast to well-established process models (e.g., Waterfall, Scrum, Crisp-DM, and the V-Model), the Process model for AI Systems Engineering (PAISE) manages data sources in individual processes of functional decomposition, quality assurance, and data provisioning. PAISE relies on rich specifications that drive an iterative implementation of functionalities with checkpoints that validate requirements satisfaction.

\textcite{nitta2022ethics} introduce an AI Ethical Impact model to close the gap between guidelines defined by institutions and the systems' ethical requirements. The model extracts the risks of AI systems based on the User Experience (UX) quality model. This model decomposes the systems' components and explicitly annotates their interactions with ethical risks.~\textcite{cody2023precepts} propose a formal model for engineering intelligent systems with diffuse boundaries between components. The authors introduce the precepts of core and periphery, which rely on set theory to represent systems structures that do not change (i.e., core) and structures that change (i.e., periphery). These precepts enable dynamic analysis of the relationships between core and peripheral structures (i.e., the \textit{System Dynamics} principle). Such analysis allows the regulation of systems to satisfy requirements.~\textcite{dzambic2022patterns} propose architectural patterns to integrate AI technologies into safety-critical systems. These patterns decompose the systems and validate the outputs of AI-based components before impacting requirements. The architectural patterns propose three alternatives (i.e., the \textit{Variant Creation} principle) to implement such validation: human-based, policy-based, or model-based.~\textcite{lavin2022technology} introduce the ML Technology Readiness Levels (MLTRL) framework that relies on well-defined processes for building systems in critical domains (e.g., the TRL framework used by NASA and DARPA). The MLTRL framework includes ten levels grouped into four stages: research, prototyping, productisation, and deployment. These levels support alignment and reliability by defining requirements from all stakeholders, exploring and experimenting with alternatives through developing proofs of concept (i.e., the \textit{Variant Creation} principle), and specifying rigorous assessment between levels.~\textcite{hershey2021acdans} proposes ACDANS, a system of systems Approach for Complex Deterministic and Non-deterministic Systems. ACDANS propose a workflow of five steps to improve the reliability, safety, and security of SoS. This workflow focuses on requirements first. These requirements drive a modelling and simulation process where different solutions are tested (i.e., the \textit{System Dynamics} and \textit{Variant Creation} principles). The combination of requirements and solutions testing supports systems alignment and reliability.

\subsection{Interpretability and Accountability}

\begin{figure}[t!]
    \centering
    \begin{tikzpicture}[scale=0.18]
        \tkzKiviatDiagram[lattice=8, label space=2.75, gap=1.125]{System Views, Agility Systems, System Dynamics, Top\\Down, Variant Creation, Problem-Solving Cycle}
        \tkzKiviatLine[mark=ball, mark size=6pt,color=blue, fill=blue, opacity=.25](6,0,2,2,0,3)
    \end{tikzpicture}
    \vspace{-4mm}
    \caption{Radar chart comparing applied principles in the context of Interpretability and Accountability.}
    \label{fig:radar-interpretability}
    \vspace{-7mm}
\end{figure}
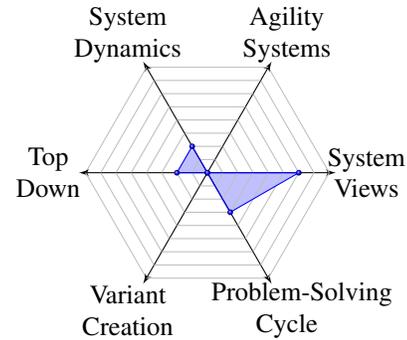

Figure~\ref{fig:radar-interpretability} shows how the selected papers approach interpretability and accountability challenges (i.e., above 30\% of works). Six of these papers apply the \textit{Systems View} principle, and none of them use the \textit{Agility Systems} and the \textit{Variant Creation} ones. Analysing problems and systems from different perspectives defines the roles of systems' actors as a base for accountability processes.~\textcite{mokander2023auditing} propose to audit large language models and their applications using three views: governance audits (i.e., organisations that develop LLMs), model audits (i.e., trained LLMs), and application audits (i.e., downstream applications that use LLMs). Different views on the auditing process facilitate the allocation of responsibility to technology providers and downstream developers when systems misbehave.~\textcite{solomonides2021amias} define a deployment phase into their governance principles and systems lifecycle to monitor socio-technical systems in healthcare. Accountable parties must use auditing mechanisms and report the results to oversight bodies in such phase. Similarly, the ML governance model proposed by~\textcite{laato2022governance} maps accountability and interpretability requirements to the system development and operation dimensions. These phases identify sources of failure by versioning and detecting anomalies in data and ML models.

The Confiance.ai framework identifies relationships between systems components by decomposing systems using ten different viewpoints. The assurance use case~\parencite{adedjouma2022dependable} shows that such explicit modelling makes systems' behaviour traceable and auditable. Making systems more transparent and accountable positively impacts the perception of healthcare systems~\parencite{wang2022health}. The authors focus on improving transparency and accountability of healthcare digital systems in governance and digital layers. The governance layer regulates the industry and establishes education policies for developers, physicians, and patients. The digital layer designs systems that can be inspected and supervised by the related party to secure patients' confidentiality. The paper proposes to use a causal graph to model the interactions between components in complex socio-technical systems and facilitate accountability analyses (i.e., the \textit{System Dynamics} principle). The MLTRL framework~\parencite{lavin2022technology} proposes using data-oriented architectures~\parencite{cabrera2023real} to support systems' accountability and interpretability. This framework defines rigorous monitoring and testing processes to identify data quality issues and data drifts at deployment. The formal method proposed by~\textcite{cody2023precepts} supports identifying which components generate a given behaviour or output at different system levels. The concepts of core and peripheral structures quantify the influence of components inside the system.

\subsection{Maintainability and Sustainability}

\begin{figure}[t!]
    \centering
    \begin{tikzpicture}[scale=0.18]
        \tkzKiviatDiagram[lattice=8, label space=2.75, gap=1.125]{System Views, Agility Systems, System Dynamics, Top\\Down, Variant Creation, Problem-Solving Cycle}
        \tkzKiviatLine[mark=ball, mark size=6pt,color=blue, fill=blue, opacity=.25](3,1,2,3,4,4)
    \end{tikzpicture}
    \vspace{-4mm}
    \caption{Radar chart comparing applied principles in the context of Maintainability and Sustainability.}
    \label{fig:radar-maintainability}
    \vspace{-7mm}
\end{figure}
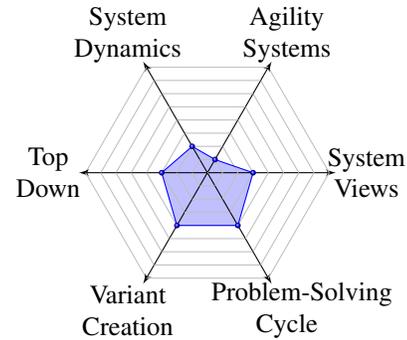

Figure~\ref{fig:radar-maintainability} shows how the works use the principles to address maintainability and sustainability issues. We found that three papers support effective resource management by using cost-benefit analysis. ACDANS~\parencite{hershey2021acdans} relies on a workflow that uses a Modelling \& Simulation (M\&S) tool that permits the evaluation of different solutions from a cost-benefit perspective. The MLTRL framework~\parencite{lavin2022technology} supports systems maintainability and sustainability in two ways. First, defining requirements and constant checkpoints reduce the risk of wasting resources. Second, evaluating alternatives enables designers to make informed decisions when selecting a solution.~\textcite{folds2019digitaltwin} propose using digital twins models for systems analysis at the design and operation stages. The digital twin models rely on Mission Function Task (MFT) analysis, which creates different design scenarios for a system. The application of the MFT analysis in the digital twin model enables dynamic modelling, simulation, and experimentation of systems, which can include, among other variables, their resource consumption.

Five papers propose approaches for systems self-maintenance. The microservices architectures proposed by~\textcite{langford2022trusted} implement the  MAPE-K loop for systems self-adaptation tasks at runtime. These tasks include a flexible updating of learning models (i.e., the \textit{Agility Systems} principle). The formal method proposed by~\textcite{cody2023precepts} allows temporal analysis between systems. Changes in the core and peripheral structures represent the evolution of systems over time to support maintenance decision-making processes from designers or the system itself. One of the two architectural patterns proposed by~\textcite{dzambic2022patterns} explores alternatives for integrating AI into autonomous architectures for safety-critical systems. These alternatives explore architectural configurations in the Cloud and at the Edge infrastructures. The system operation phase of the ML governance model introduced by~\textcite{laato2022governance} focuses on systems' testing and maintenance. The goal is to support failure identification and recovery when systems are deployed and running. Similarly, the governance principles defined by~\textcite{solomonides2021amias} consider systems' constant monitoring and maintenance, including updating tasks that enable systems to respond to changing requirements.

\subsection{Security and Privacy}

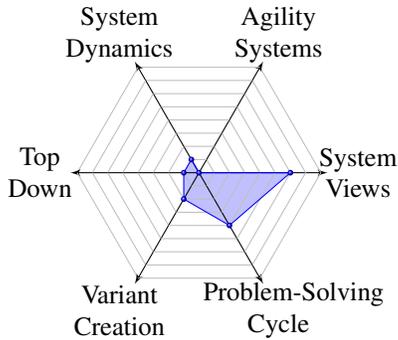
\begin{figure}[t!]
    \centering
    \begin{tikzpicture}[scale=0.18]
        \tkzKiviatDiagram[lattice=8, label space=2.75, gap=1.125]{System Views, Agility Systems, System Dynamics, Top\\Down, Variant Creation, Problem-Solving Cycle}
        \tkzKiviatLine[mark=ball, mark size=6pt,color=blue, fill=blue, opacity=.25](6,0,1,1,2,4)
    \end{tikzpicture}
    \vspace{-4mm}
    \caption{Radar chart comparing applied principles in the context of Security and Privacy.}
    \label{fig:radar-security}
    \vspace{-7mm}
\end{figure}

Eight of the twenty-four selected papers propose approaches to address security and privacy issues (Figure~\ref{fig:radar-security}). The most used principles are the \textit{System Views} and the \textit{Problem-Solving Cycle}. None of the papers apply the \textit{Agility Systems} principle.~\textcite{cai2020safetyai} proposes a checklist for analysing AI systems safety from different perspectives. The paper proposes decomposing autonomous systems into parts to apply the safety checklist. These parts represent potential vulnerabilities of the system, including system components, structures, assumptions, properties, and interfaces. The safety checklist assesses systems sensors, data collection tasks, default design assumptions, systems observability and controllability capabilities, the complexity of the systems, and their interactions with humans. Similarly,~\textcite{fujii2020guidelines} establish checkpoints along the whole AI system lifecycle to represent different analysis viewpoints. The data integrity, system quality, and customer expectation viewpoints represent security and privacy requirements for assuring the quality of ML-based systems.~\textcite{mokander2023auditing} defines that security and privacy issues caused by LLMs can emerge from technology providers or downstream applications. The authors structure the audit process into three layers to assign responsibilities and identify and prevent security and privacy risks. The causal graph proposed by~\textcite{wang2022health} facilitates modelling relationships between patients, physicians, AI components, and governance institutions in complex socio-technical systems. This graph is a formal representation that enables multiple analyses (e.g., identifying security and privacy vulnerabilities). ~\textcite{solomonides2021amias} propose principles for adopting AI into healthcare systems. These principles include data management tasks, auditing mechanisms, systems decommissioning, and external standards to assure patients' data privacy. For example, the systems decommissioning principles deal with the closure, curation, and maintenance of records of AI systems. Institutions must preserve data for medical and legal reasons (e.g., hospitals must keep data records from systems in the paediatrics domain for 21 years). Likewise, the development dimension of the ML governance model proposed by~\textcite{laato2022governance} maps the data privacy and validation requirements to the systems development process. This mapping makes particular emphasis on the ethical use of sensible data. The MLTR framework~\parencite{lavin2022technology} includes data intervention tests and monitoring at integration and deployment levels. These tests assess possible security and privacy issues and solution alternatives. The Modelling \& Simulation tool of ACDANS~\parencite{hershey2021acdans} simulates and evaluates different systems' scenarios. These scenarios can include security vulnerabilities and their possible solutions.
\section{Open Research Directions}
\label{sec:open-directions}

Our survey shows how current research uses systems approaches to mitigate the challenges that emerge from adopting AI. We discuss open research challenges based on these findings. 

\subsection{Inclusive Requirements Definition}

The \textit{Systems Views and Top Down} principles enable considering different perspectives and analysis levels when defining requirements according to our survey. However, these principles work well when the stakeholders are inside the same organisation and have a common language. Socio-technical systems involve heterogeneous subsystems that impact large populations of the public. Defining comprehensive requirements for such systems is an open challenge because these can impact multiple stakeholders who do not communicate with each other. Systems engineering methodologies must consider public engagement activities to get an inclusive problem definition. Educational policies are crucial as the public can misunderstand the implications of technologies. This misunderstanding creates both fake hopes and fears, which negatively impacts the LLMs' adoption. A promising direction to address these challenges is designing AI systems driven by public value~\parencite{bastidas2024aipublic}. In addition, new AI-based interfaces can empower people when defining ML-based systems~\parencite{robinson2024requirements}.

\subsection{Operationalising High-Level Concepts}

The surveyed papers mostly rely on the \textit{Top Down} principle to translate requirements into artefacts that satisfy them using systems' design metamodels, architectures, and methodologies. However, socio-technical systems requirements are hard to operationalise. There is a gap between high-level concepts like sustainability, robustness, and truthfulness and their technical implementations (e.g., LLMs' fairness metrics). Reductionist techniques aim to close this gap. However, systems based on such techniques usually fail to capture the real-world complexity. AI-based socio-technical systems require novel approaches to manage the interaction between multiple actors, subsystems, and data sources during the system's lifecycle. The \textit{System Dynamics} and \textit{Agility Systems} principles require more attention as they can provide dynamic and flexible tools~\parencite{folds2019digitaltwin,yu2024management} and architectures~\parencite{cabrera2023real} for systems operation~\parencite{cabrera2024self}.

\subsection{Changing organisations’ culture}

Most surveyed papers support safety-critical systems. The \textit{Problem-Solving Cycle} principle defines development methodologies for such systems, which prioritise the documentation, research, planning, and prototyping stages before moving to implementation. However, current organisations profess an agile culture where working software is the measure of progress~\parencite{srivastava2017scrum}. Including AI-based components that designers do not fully understand lowers the bar to qualify a system as safety-critical. Organisations need a shift towards methodologies that balance or even prioritise planning phases against production phases. Current methods for designing critical systems provide good directions~\parencite{lavin2022technology}. Automating the systems implementation stages (e.g., code generation) frees resources and leverages such methodological shifts~\parencite{robinson2024requirements}.

\subsection{Dynamic Technological Landscape}

The technological ecosystem around LLMs is changing quickly. New learning models and tools challenge the completeness and relevance of systems guidelines, metamodels, methodologies, architectural patterns, and formal methods that assume or require complete knowledge about the system actors, enablers, benefits, and risks. The most applied principles (i.e., \textit{Systems Views, Top Down, and Problem-Solving Cycle}) are static and rely on prior knowledge. We must research how to combine these with the dynamic and flexible ones (i.e., \textit{Agility Systems, System Dynamics, and Variant Creation}). Neurosymbolic approaches and ML for systems engineering are promising directions in this regard~\parencite{decardi2024generative}.
\section{Conclusions}
\label{sec:conclusions}

This paper surveys systems research efforts for engineering AI-based systems. We show the systems engineering approach offers a good starting point for addressing the LLMs' challenges. Most works prioritise addressing the systems \textit{Alignment and Reliability} issues. The systems engineering principles that guide most of the approaches are the \textit{Systems Views}, the \textit{Top-Down}, and the \textit{Problem-Solving Cycle}. The \textit{Agility Systems}, \textit{System Dynamics}, and \textit{Variant Creation} principles require more attention. We plan to use the outputs of this survey to provide systems engineering principles and guidance for designing and developing AI-based systems integrating LLMs. These principles will drive our work towards building an ecosystem of architectural patterns, design artefacts, and software platforms to develop and deploy sustainable AI-based systems.
\printbibliography

@article{feuerriegel2024generative,
  title={Generative AI},
  author={Feuerriegel, Stefan and Hartmann, Jochen and Janiesch, Christian and Zschech, Patrick},
  journal={Business \& Information Systems Engineering},
  volume={66},
  number={1},
  pages={111--126},
  year={2024},
  publisher={Springer}
}

@article{cabrera2023real,
  title={Real-world Machine Learning Systems: A survey from a Data-Oriented Architecture Perspective},
  author={Cabrera, Christian and Paleyes, Andrei and Thodoroff, Pierre and Lawrence, Neil D},
  journal={arXiv preprint arXiv:2302.04810},
  year={2023}
}

@article{schwartz2020greenai,
  author = {Schwartz, Roy and Dodge, Jesse and Smith, Noah A. and Etzioni, Oren},
  title = {Green AI},
  year = {2020},
  issue_date = {December 2020},
  publisher = {Association for Computing Machinery},
  address = {New York, NY, USA},
  volume = {63},
  number = {12},
  issn = {0001-0782},
  url = {https://doi.org/10.1145/3381831},
  doi = {10.1145/3381831},
  journal = {Commun. ACM},
  month = {11},
  pages = {54–63},
  numpages = {10}
}

@inproceedings{cabrera2024self,
  title={Self-sustaining Software Systems (S4): Towards Improved Interpretability and Adaptation},
  author={Cabrera, Christian and Paleyes, Andrei and Lawrence, Neil D},
  booktitle={2024 International Workshop New Trends in Software Architecture (SATrends’24)},
  year={2024},
  publisher={Association for Computing Machinery},
  address = {New York, NY, USA},
  doi = {10.1145/3643657.3643910},
  month = {4},
  pages={5}
}

@article{vanderlinde2022aisechallenges,
  author = {Jake Vanderlinde, Kevin Robinson and Benjamin Mashford},
  title = {The challenges for artificial intelligence and systems engineering},
  journal = {Australian Journal of Multi-Disciplinary Engineering},
  volume = {18},
  number = {1},
  pages = {47--53},
  year = {2022},
  publisher = {Taylor \& Francis},
  doi = {10.1080/14488388.2022.2044607},
  URL = {https://doi.org/10.1080/14488388.2022.2044607},
  eprint = {https://doi.org/10.1080/14488388.2022.2044607}
}

@article{decardi2024generative,
  title={Generative AI and Process Systems Engineering: The Next Frontier},
  author={Decardi-Nelson, Benjamin and Alshehri, Abdulelah S and Ajagekar, Akshay and You, Fengqi},
  journal={arXiv preprint arXiv:2402.10977},
  year={2024}
}

@article{zhao2024expllmsurvey,
  author = {Zhao, Haiyan and Chen, Hanjie and Yang, Fan and Liu, Ninghao and Deng, Huiqi and Cai, Hengyi and Wang, Shuaiqiang and Yin, Dawei and Du, Mengnan},
  title = {Explainability for Large Language Models: A Survey},
  year = {2024},
  issue_date = {April 2024},
  publisher = {Association for Computing Machinery},
  address = {New York, NY, USA},
  volume = {15},
  number = {2},
  issn = {2157-6904},
  url = {https://doi.org/10.1145/3639372},
  doi = {10.1145/3639372},
  journal = {ACM Trans. Intell. Syst. Technol.},
  month = {2},
  articleno = {20},
  numpages = {38}
}

@article{dantonoli2024large,
  title={Large language models in radiology: fundamentals, applications, ethical considerations, risks, and future directions},
  author={D’Antonoli, Tugba Akinci and Stanzione, Arnaldo and Bluethgen, Christian and Vernuccio, Federica and Ugga, Lorenzo and Klontzas, Michail E and Cuocolo, Renato and Cannella, Roberto and Ko{\c{c}}ak, Burak},
  journal={Diagnostic and Interventional Radiology},
  volume={30},
  number={2},
  pages={80},
  year={2024},
  publisher={Turkish Society of Radiology}
}

@article{sommerville2019artificial,
  title={Artificial intelligence and systems engineering},
  author={Sommerville, Ian},
  journal={Prospects for Artificial Intelligence: Proceedings of AISB},
  volume={93},
  pages={29},
  year={2019}
}

@inbook{llinas2021review,
  author="Llinas, James
  and Fouad, Hesham
  and Mittu, Ranjeev",
  editor="Lawless, William F.
  and Mittu, Ranjeev
  and Sofge, Donald A.
  and Shortell, Thomas
  and McDermott, Thomas A.",
  title="Systems Engineering for Artificial Intelligence-based Systems: A Review in Time",
  bookTitle="Systems  Engineering and Artificial Intelligence ",
  year="2021",
  publisher="Springer International Publishing",
  address="Cham",
  pages="93--113",
  isbn="978-3-030-77283-3",
  doi="10.1007/978-3-030-77283-3_6",
  url="https://doi.org/10.1007/978-3-030-77283-3_6"
}

@article{wade2020vlsirev,
  author = {Wade, Jon and Buenfil, Jorge and Collopy, Paul},
  title = {A Systems Engineering Approach for Artificial Intelligence: Inspired by the VLSI Revolution of Mead \& Conway},
  journal = {INSIGHT},
  volume = {23},
  number = {1},
  pages = {41-47},
  doi = {https://doi.org/10.1002/inst.12284},
  url = {https://incose.onlinelibrary.wiley.com/doi/abs/10.1002/inst.12284},
  eprint = {https://incose.onlinelibrary.wiley.com/doi/pdf/10.1002/inst.12284},
  year = {2020}
}

@article{pfrommer2022ki,
  title={KI-Engineering--AI Systems Engineering: Systematic development of AI as part of systems that master complex tasks},
  author={Pfrommer, Julius and Usl{\"a}nder, Thomas and Beyerer, J{\"u}rgen},
  journal={at-Automatisierungstechnik},
  volume={70},
  number={9},
  pages={756--766},
  year={2022},
  publisher={De Gruyter Oldenbourg}
}

@article{baxter2010socio,
  author = {Baxter, Gordon and Sommerville, Ian},
  title = "{Socio-technical systems: From design methods to systems engineering}",
  journal = {Interacting with Computers},
  volume = {23},
  number = {1},
  pages = {4-17},
  year = {2010},
  month = {08},
  issn = {0953-5438},
  doi = {10.1016/j.intcom.2010.07.003},
  url = {https://doi.org/10.1016/j.intcom.2010.07.003},
  eprint = {https://academic.oup.com/iwc/article-pdf/23/1/4/2038336/iwc23-0004.pdf}
}

@article{bastidas2022concepts,
  title={Concepts for modeling smart cities: An archimate extension},
  author={Bastidas, Viviana and Reychav, Iris and Ofir, Alon and Bezbradica, Marija and Helfert, Markus},
  journal={Business \& Information Systems Engineering},
  volume={64},
  number={3},
  pages={359--373},
  year={2022},
  publisher={Springer}
}

@inbook{zittrain2022debt, 
  place={Cambridge}, 
  series={Cambridge Law Handbooks}, 
  title={Intellectual Debt: With Great Power Comes Great Ignorance}, 
  booktitle={The Cambridge Handbook of Responsible Artificial Intelligence: Interdisciplinary Perspectives}, publisher={Cambridge University Press}, 
  author={Zittrain, Jonathan}, 
  editor={Voeneky, Silja and Kellmeyer, Philipp and Mueller, Oliver and Burgard, WolframEditors}, 
  year={2022}, 
  pages={176–184}, 
  collection={Cambridge Law Handbooks}
}

@article{kitchenham20132049,
   title = {A systematic review of systematic review process research in software engineering},
   journal = {Information and Software Technology},
   volume = {55},
   number = {12},
   pages = {2049-2075},
   year = {2013},
   issn = {0950-5849},
   doi = {https://doi.org/10.1016/j.infsof.2013.07.010},
   url = {https://www.sciencedirect.com/science/article/pii/S0950584913001560},
   author = {Barbara Kitchenham and Pearl Brereton}
}

@book{haberfellner2019systems,
  title={Systems engineering: fundamentals and applications},
  author={Haberfellner, Reinhard and De Weck, Olivier and Fricke, Ernst and V{\"o}ssner, Siegfried},
  year={2019},
  publisher={Springer}
}

@article{mokander2023auditing,
  title={Auditing large language models: a three-layered approach},
  author={M{\"o}kander, Jakob and Schuett, Jonas and Kirk, Hannah Rose and Floridi, Luciano},
  journal={AI and Ethics},
  pages={1--31},
  year={2023},
  publisher={Springer}
}

@inproceedings{meyer2019seagents,
  author = {Meyer, Ole and Gruhn, Volker},
  title = {Towards concept based software engineering for intelligent agents},
  year = {2019},
  publisher = {IEEE Press},
  url = {https://doi.org/10.1109/RAISE.2019.00015},
  doi = {10.1109/RAISE.2019.00015},
  booktitle = {Proceedings of the 7th International Workshop on Realizing Artificial Intelligence Synergies in Software Engineering},
  pages = {42–48},
  numpages = {7},
  location = {Montreal, Quebec, Canada},
  series = {RAISE '19}
}

@article{cai2020safetyai,
  author = {Cai, Yang},
  title = {Safety Analytics for AI Systems},
  year = {2020},
  journal = {Lecture Notes in Computer Science (including subseries Lecture Notes in Artificial Intelligence and Lecture Notes in Bioinformatics)},
  volume = {12424 LNCS},
  pages = {434 – 448},
  doi = {10.1007/978-3-030-60117-1_32},
  url = {https://www.scopus.com/inward/record.uri?eid=2-s2.0-85094144617&doi=10.1007%2f978-3-030-60117-1_32&partnerID=40&md5=912d3292340e6ae22484bc2f072645db},
  type = {Conference paper},
  publication_stage = {Final},
  source = {Scopus},
  note = {Cited by: 0}
}

@article{hasterok2022process,
  url = {https://doi.org/10.1515/auto-2022-0020},
  title = {PAISE® – process model for AI systems engineering},
  author = {Constanze Hasterok and Janina Stompe},
  pages = {777--786},
  volume = {70},
  number = {9},
  journal = {at - Automatisierungstechnik},
  doi = {doi:10.1515/auto-2022-0020},
  year = {2022},
  lastchecked = {2024-05-22}
}

@inproceedings{wang2022health,
  author={Wang, Bijun and Asan, Onur and Mansouri, Mo},
  booktitle={2022 IEEE International Symposium on Systems Engineering (ISSE)}, 
  title={Patients’ Perceptions of Integrating AI into Healthcare: Systems Thinking Approach}, 
  year={2022},
  pages={1-6},
  doi={10.1109/ISSE54508.2022.10005383}
}

@inproceedings{mattioli2023industry,
  author={Mattioli, J. and Roux, X. Le and Braunschweig, B. and Cantat, L. and Tschirhart, F. and Robert, B. and Gelin, R. and Nicolas, Y.},
  booktitle={2023 Fifth International Conference on Transdisciplinary AI (TransAI)}, 
  title={AI Engineering to Deploy Reliable AI in Industry}, 
  year={2023},
  volume={},
  number={},
  pages={228-231},
  doi={10.1109/TransAI60598.2023.00015}
}

@article{solomonides2021amias,
  author = {Solomonides, Anthony E and Koski, Eileen and Atabaki, Shireen M and Weinberg, Scott and McGreevey, John D, III and Kannry, Joseph L and Petersen, Carolyn and Lehmann, Christoph U},
  title = "{Defining AMIA’s artificial intelligence principles}",
  journal = {Journal of the American Medical Informatics Association},
  volume = {29},
  number = {4},
  pages = {585-591},
  year = {2021},
  month = {11},
  issn = {1527-974X},
  doi = {10.1093/jamia/ocac006},
  url = {https://doi.org/10.1093/jamia/ocac006},
  eprint = {https://academic.oup.com/jamia/article-pdf/29/4/585/42897428/ocac006.pdf}
}

@inproceedings{lavazza2021intensive,
  author={Lavazza, Luigi and Morasca, Sandro},
  booktitle={2021 IEEE/ACM 1st Workshop on AI Engineering - Software Engineering for AI (WAIN)}, 
  title={Understanding and Modeling AI-Intensive System Development}, 
  year={2021},
  pages={55-61},
  doi={10.1109/WAIN52551.2021.00015}
}

@article{fujii2020guidelines,
  title={Guidelines for quality assurance of machine learning-based artificial intelligence},
  author={Fujii, Gaku and Hamada, Koichi and Ishikawa, Fuyuki and Masuda, Satoshi and Matsuya, Mineo and Myojin, Tomoyuki and Nishi, Yasuharu and Ogawa, Hideto and Toku, Takahiro and Tokumoto, Susumu and others},
  journal={International journal of software engineering and knowledge engineering},
  volume={30},
  number={11n12},
  pages={1589--1606},
  year={2020},
  publisher={World Scientific}
}

@article{lavin2022technology,
  title={Technology readiness levels for machine learning systems},
  author={Lavin, Alexander and Gilligan-Lee, Ciar{\'a}n M and Visnjic, Alessya and Ganju, Siddha and Newman, Dava and Ganguly, Sujoy and Lange, Danny and Baydin, At{\'\i}l{\'\i}m G{\"u}ne{\c{s}} and Sharma, Amit and Gibson, Adam and others},
  journal={Nature Communications},
  volume={13},
  number={1},
  pages={6039},
  year={2022},
  publisher={Nature Publishing Group UK London}
}

@inproceedings{pedroza2019safe,
  title={Safe-by-design development method for artificial intelligent based systems},
  author={Pedroza, Gabriel and Adedjouma, Morayo},
  booktitle={SEKE 2019: The 31st International Conference on Software Engineering and Knowledge Engineering},
  pages={391--397},
  year={2019}
}

@inproceedings{hershey2021acdans,
  author={Hershey, Paul},
  booktitle={2021 16th International Conference of System of Systems Engineering (SoSE)}, 
  title={System of Systems Engineering Approach for Complex Deterministic and Nondeterministic Systems (ACDANS)}, 
  year={2021},
  volume={},
  number={},
  pages={185-190},
  doi={10.1109/SOSE52739.2021.9497496}
}

@article{yu2024management,
  author={Yu, Youshan and Lakemond, Nicolette and Holmberg, Gunnar},
  journal={IEEE Transactions on Engineering Management}, 
  title={AI in the Context of Complex Intelligent Systems: Engineering Management Consequences}, 
  year={2024},
  volume={71},
  number={},
  pages={6512-6525},
  doi={10.1109/TEM.2023.3268340}
}

@inproceedings{folds2019digitaltwin,
  author={Folds, Dennis J. and McDermott, Thomas A.},
  booktitle={2019 IEEE International Conference on Systems, Man and Cybernetics (SMC)}, 
  title={The Digital (Mission) Twin: an Integrating Concept for Future Adaptive Cyber-Physical-Human Systems}, 
  year={2019},
  volume={},
  number={},
  pages={748-754},
  doi={10.1109/SMC.2019.8914324}
}

@inproceedings{langford2022trusted,
  author={Langford, Michael Austin and Cheng, Betty H.C.},
  booktitle={2022 IEEE International Conference on Autonomic Computing and Self-Organizing Systems (ACSOS)}, 
  title={A Modular and Composable Approach to Develop Trusted Artificial Intelligence}, 
  year={2022},
  volume={},
  number={},
  pages={121-130},
  doi={10.1109/ACSOS55765.2022.00030}
}

@inproceedings{cody2023precepts,
  author="Cody, Tyler
  and Shadab, Niloofar
  and Salado, Alejandro
  and Beling, Peter",
  editor="Goertzel, Ben
  and Ikl{\'e}, Matt
  and Potapov, Alexey
  and Ponomaryov, Denis",
  title="Core and Periphery as Closed-System Precepts for Engineering General Intelligence",
  booktitle="Artificial General Intelligence",
  year="2023",
  publisher="Springer International Publishing",
  address="Cham",
  pages="209--219",
  isbn="978-3-031-19907-3"
}

@inproceedings{nitta2022ethics,
  author={Nitta, Izumi and Ohashi, Kyoko and Shiga, Satoko and Onodera, Sachiko},
  booktitle={2022 IEEE 30th International Requirements Engineering Conference Workshops (REW)}, 
  title={AI Ethics Impact Assessment based on Requirement Engineering}, 
  year={2022},
  volume={},
  number={},
  pages={152-161}
}

@inproceedings{dzambic2022patterns,
  author = {Dzambic, Maid and Dobaj, J\"{u}rgen and Seidl, Matthias and Macher, Georg},
  title = {Architectural Patterns for Integrating AI Technology into Safety-Critical Systems},
  year = {2022},
  isbn = {9781450389976},
  publisher = {Association for Computing Machinery},
  address = {New York, NY, USA},
  url = {https://doi.org/10.1145/3489449.3490014},
  doi = {10.1145/3489449.3490014},
  booktitle = {Proceedings of the 26th European Conference on Pattern Languages of Programs},
  articleno = {36},
  numpages = {8},
  location = {<conf-loc>, <city>Graz</city>, <country>Austria</country>, </conf-loc>},
  series = {EuroPLoP '21}
}

@article{salwei2022healthcare,
  author={Megan E. Salwei and Pascale Carayon},
  title={A Sociotechnical Systems Framework for the Application of Artificial Intelligence in Health Care Delivery},
  journal={Journal of Cognitive Engineering and Decision Making},
  volume={16},
  number={4},
  pages={194-206},
  year={2022},
  doi={10.1177/15553434221097357},
  note={PMID: 36704421},
  url={https://doi.org/10.1177/15553434221097357},
  eprint={https://doi.org/10.1177/15553434221097357}
}

@inproceedings{adedjouma2022dependable,
  author={Adedjouma, Morayo and Alix, Christophe and Cantat, Loic and Jenn, Eric and Mattioli, Juliette and Robert, Boris and Tschirhart, Fabien and Voirin, Jean-Luc},
  booktitle={2022 17th Annual System of Systems Engineering Conference (SOSE)}, 
  title={Engineering Dependable AI Systems}, 
  year={2022},
  pages={458-463},
  doi={10.1109/SOSE55472.2022.9812672}
}

@inproceedings{laato2022governance,
  author={Laato, Samuli and Birkstedt, Teemu and M\"{a}antym\"{a}ki, Matti and Minkkinen, Matti and Mikkonen, Tommi},
  title={AI governance in the system development life cycle: insights on responsible machine learning engineering},
  year={2022},
  isbn={9781450392754},
  publisher={Association for Computing Machinery},
  address={New York, NY, USA},
  url={https://doi.org/10.1145/3522664.3528598},
  doi={10.1145/3522664.3528598},
  booktitle={Proceedings of the 1st International Conference on AI Engineering: Software Engineering for AI},
  pages={113–123},
  numpages={11},
  location={Pittsburgh, Pennsylvania},
  series={CAIN '22}
}

@article{zeller2024toward,
  title={Toward a safe MLOps process for the continuous development and safety assurance of ML-based systems in the railway domain},
  author={Zeller, Marc and Waschulzik, Thomas and Schmid, Reiner and Bahlmann, Claus},
  journal={AI and Ethics},
  pages={1--8},
  year={2024},
  publisher={Springer}
}

@article{nabavi2022five,
  title={Five Ps: Leverage zones towards responsible AI},
  author={Nabavi, Ehsan and Browne, Chris},
  journal={arXiv preprint arXiv:2205.01070},
  year={2022}
}

@article{hedin2023strengthening,
  title={Strengthening the AI Operating Environment},
  author={Hedin, Bruce and Curtis, Samuel},
  year={2023}
}

@inproceedings{srivastava2017scrum,
  title={SCRUM model for agile methodology},
  author={Srivastava, Apoorva and Bhardwaj, Sukriti and Saraswat, Shipra},
  booktitle={2017 International Conference on Computing, Communication and Automation (ICCCA)},
  pages={864--869},
  year={2017},
  organization={IEEE}
}

@article{robinson2024requirements,
  title={Requirements are All You Need: The Final Frontier for End-User Software Engineering},
  author={Robinson, Diana and Cabrera, Christian and Gordon, Andrew D and Lawrence, Neil D and Mennen, Lars},
  journal={arXiv preprint arXiv:2405.13708},
  year={2024}
}

@article{bastidas2024aipublic,
  title={Socio-Technical {AI} Design For Public Value},
  journal={AIS Electronic Library (AISeL)},
  author={Bastidas, Viviana and Schooling, Jennifer},
  year={2024},
  url={https://aisel.aisnet.org/treos_ecis2024/78}
}
\end{document}